\pdfoutput=1

\documentclass[11pt]{article}

\usepackage{amsmath,amsfonts,bm}

\def\eqref#1{equation~(\ref{#1})}

\def\1{\bm{1}}

\DeclareMathAlphabet{\mathsfit}{\encodingdefault}{\sfdefault}{m}{sl}
\SetMathAlphabet{\mathsfit}{bold}{\encodingdefault}{\sfdefault}{bx}{n}

\usepackage{graphicx}
\usepackage{amsmath,amssymb}
\usepackage{mathtools}
\usepackage{amsthm}
\usepackage{bbm}
\usepackage{float}
\usepackage{accents}
\usepackage{hyperref}
\usepackage[]{algorithm}
\usepackage[noend]{algorithmic}
\usepackage{wrapfig}
\usepackage[utf8]{inputenc} %
\usepackage[T1]{fontenc}    %
\usepackage{url}            %
\usepackage{booktabs}       %
\usepackage{amsfonts}       %
\usepackage{nicefrac}       %
\usepackage{microtype}      %
\usepackage{xcolor}         %
\usepackage{enumitem}
\usepackage{caption}
\usepackage{subcaption}
\usepackage{graphicx}
\usepackage{epstopdf,epsfig}
    \theoremstyle{plain}
    \newtheorem{theorem}{Theorem}[section]
    
    \newtheorem{lemma}[theorem]{Lemma}
    
    \theoremstyle{definition}
    \newtheorem{assumption}[theorem]{Assumption}
    \theoremstyle{remark}
    \newtheorem{remark}[theorem]{Remark}

\def\remark{\addtocounter{remark}{1}\def\@currentlabel{\theremark}%
\emph{Remark~\theremark}. } \makeatother
\newcounter{remark}

\usepackage[suppress]{color-edits}
 \addauthor{mh}{red}
 \addauthor{xw}{magenta}
 
\usepackage[]{acl}

\usepackage{multirow}
\usepackage{times}
\usepackage{latexsym}
\usepackage{soul}
\usepackage{url}
\usepackage{bm}
\usepackage{hyperref}
\usepackage[utf8]{inputenc}
\usepackage{graphicx}
\usepackage{amsmath}
\usepackage{mathrsfs}
\usepackage{amsthm}
\usepackage{booktabs}
\usepackage{algorithm}
\usepackage{algorithmic}
\usepackage{geometry}
\usepackage{booktabs}
\usepackage{amssymb}
\usepackage{color}
\usepackage{multicol}
\usepackage{array}
\usepackage{colortbl}
\usepackage{makecell}
\usepackage{mathtools}                
\usepackage{hyperref}                 

\graphicspath{{Figure/}}
\usepackage[T1]{fontenc}

\usepackage[utf8]{inputenc}

\usepackage{microtype}

\title{Hi-ZFO: Hierarchical Zeroth- and First-Order LLM Fine-Tuning via Importance-Guided Tensor Selection}

\author{Feihu Jin$^{1,2}$ \and Ying Tan$^{1,2,3}$ \thanks{~ Corresponding Author}\\
  $^1$ School of Intelligence Science and Technology, Peking University\\
  $^2$ Institute for Artificial Intellignce, Peking University \\
  $^3$ State Key Laboratory of General Artificial Intelligence\\ 
\text{fhjin@stu.pku.edu.cn}, \text{ytan@pku.edu.cn}}

\begin{document}
\maketitle
\begin{abstract}
Fine-tuning large language models (LLMs) using standard first-order (FO) optimization often drives training toward sharp, poorly generalizing minima. Conversely, zeroth-order (ZO) methods offer stronger exploratory behavior without relying on explicit gradients, yet suffer from slow convergence. More critically, our analysis reveals that in generative tasks, the vast output and search space significantly amplify estimation variance, rendering ZO methods both noisy and inefficient. 
To address these challenges, we propose \textbf{Hi-ZFO} (\textbf{Hi}erarchical \textbf{Z}eroth- and \textbf{F}irst-\textbf{O}rder optimization), a hybrid framework designed to synergize the precision of FO gradients with the exploratory capability of ZO estimation. Hi-ZFO adaptively partitions the model through layer-wise importance profiling, applying precise FO updates to critical layers while leveraging ZO optimization for less sensitive ones. 
Notably, ZO in Hi-ZFO is not merely a memory-saving surrogate; it is intentionally introduced as a source of "beneficial stochasticity" to help the model escape the local minima where pure FO optimization tends to stagnate. Validated across diverse generative, mathematical, and code reasoning tasks, Hi-ZFO consistently achieves superior performance while significantly reducing the training time. These results demonstrate the effectiveness of hierarchical hybrid optimization for LLM fine-tuning.
\end{abstract}
\section{Introduction}\label{sec:intro}
\begin{figure}[t]
    \centering
    \includegraphics[width=0.8\linewidth]{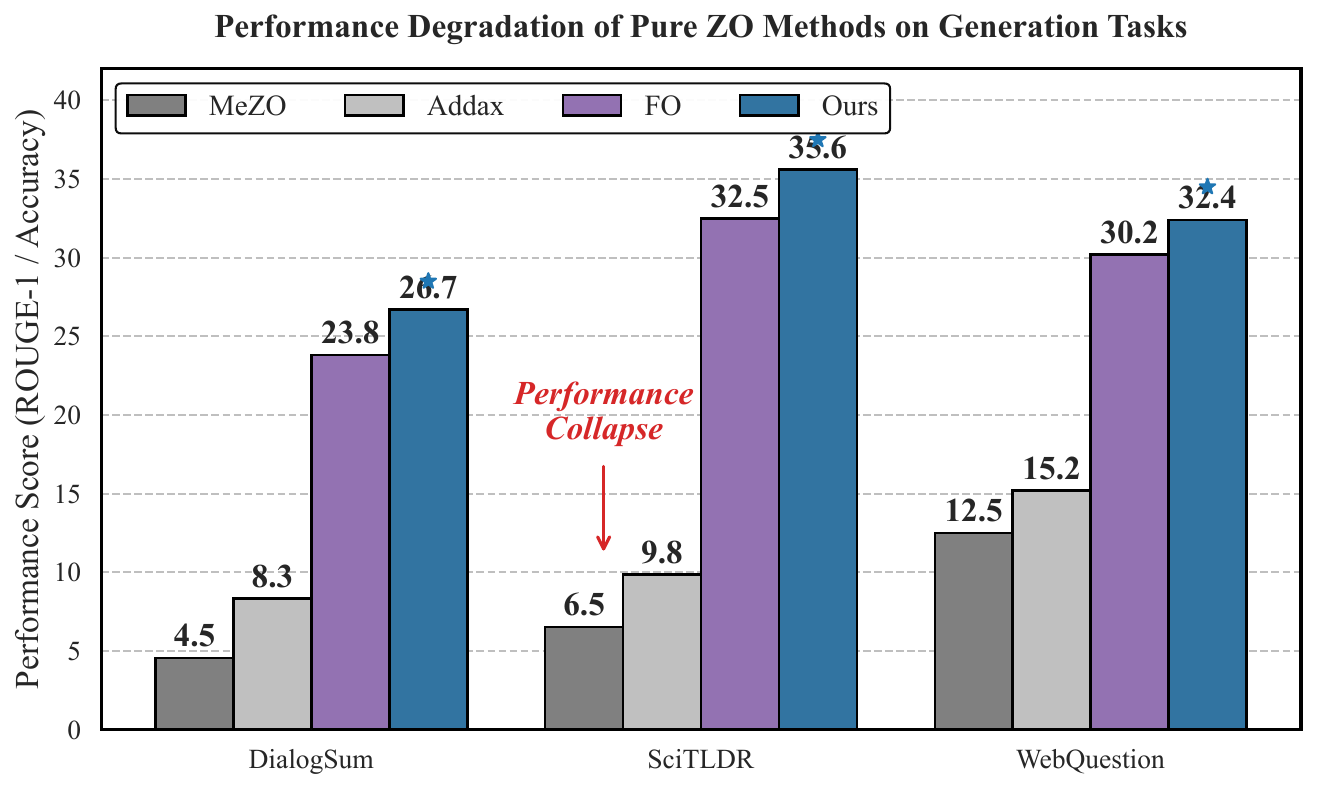}
    \caption{Performance degradation of pure ZO optimization on generation tasks. ZO methods (MeZO) exhibit a pronounced performance collapse as the output and search space expand, while Hi-ZFO remains stable and achieves substantially higher performance.
    }
    \label{fig:zo_collapse_motivation}
\end{figure}
Fine-tuning has become the prevailing paradigm for adapting large language models to specialized downstream tasks \cite{qwen3, qwen2.5, OpenAI2023GPT4TR}. The current standard relies on first-order (FO) gradient-based optimizers such as Adam \cite{DBLP:journals/corr/KingmaB14}, which are favored for their rapid convergence and precise parameter updates. Despite their success, these deterministic methods face fundamental limitations within high-dimensional, non-convex loss landscapes. Specifically, they tend to converge toward local minima and often lack the exploratory capacity required to escape such suboptimal regions \cite{keskar2016large}. Convergence to these local optima results in poor generalization capability and high model instability.

Zeroth-order (ZO) optimization \cite{Spall1992MultivariateSA} presents a compelling alternative to traditional approaches by estimating gradients via function-value perturbations instead of analytical backpropagation. By bypassing explicit gradients, ZO methods introduce a degree of inherent stochasticity that facilitates broader exploration of the parameter space, potentially allowing the optimizer to escape the local minima that often entrap first-order (FO) methods. However, the effectiveness of pure ZO optimization is constrained by the curse of dimensionality. Because the variance of gradient estimators scales linearly with the number of parameters \cite{duchi2015optimal}, these methods frequently suffer from slow convergence and instability when applied to large language models.
Our empirical analysis reveals a critical limitation, namely that ZO optimization deteriorates significantly as the complexity of the output space increases. While pure ZO methods like MeZO \cite{malladi2023mezo} and hybrid strategies such as Addax \cite{li2025addax}, which applies first-order (FO) gradients to long sequences and ZO to short ones, perform reasonably on simple classification tasks, they experience a pronounced performance collapse in generative settings (see Fig.~\ref{fig:zo_collapse_motivation}). In these scenarios, the vast search space of token sequences significantly amplifies estimation variance. This noise renders ZO updates both inefficient and erratic, eventually driving the optimization toward suboptimal regions. Consequently, a fundamental challenge arises because FO methods provide precision but lack sufficient exploration, while ZO methods encourage exploration but fail to maintain stability and efficiency in complex generative tasks.

To bridge this gap, we propose Hi-ZFO (Hierarchical Zeroth- and First-Order optimization), which is a hybrid framework designed to synergize the precision of FO with the exploration of ZO. Rather than viewing them as competing alternatives, Hi-ZFO leverages their complementarity by adopting the Importance-Guided Tensor Selection strategy from \cite{huang2024towards}. Specifically, building on this paradigm, we formulate parameter partitioning as a cost-aware optimization problem and utilize Dynamic Programming (DP) to identify the parameter subset that maximizes sensitivity importance within a fixed computational budget.

The hierarchical design yields a principled allocation of optimization regimes, whereby high-importance tensors, typically located in upper layers, receive precise FO updates to ensure stability. Meanwhile, the computation-heavy bottom layers, which are traditionally frozen to save costs, are activated via ZO optimization. By replacing static freezing with stochastic ZO adaptation, Hi-ZFO transforms these layers into a source of \textit{"beneficial stochasticity."} This mechanism serves a dual purpose, as it drastically reduces memory overhead by avoiding full backpropagation while simultaneously perturbing the optimization trajectory enough to escape the local minima where pure FO methods often stagnate. Consequently, Hi-ZFO achieves a rare synergy by combining enhanced generalization through improved landscape exploration with significant reductions in memory footprint and training time. We summarize our primary contributions as follows.
\begin{itemize}
\item We identify the cause of ZO optimization failure in generative tasks, attributing it to the high estimation variance that is significantly amplified by the expansive output search space.
\item We propose Hi-ZFO, a hierarchical framework that integrates an importance-based selection mechanism to strategically balance FO and ZO updates, thereby synergizing gradient precision with beneficial stochastic exploration.
\item Extensive evaluations demonstrate that Hi-ZFO consistently outperforms competitive baselines while simultaneously achieving substantial reductions in memory overhead and accelerating convergence.
\end{itemize}

\section{Related Work}
\noindent\textbf{Zeroth-Order and Hybrid Optimization.}
Zeroth-order (ZO) methods like MeZO \citep{malladi2023mezo} enable memory-efficient fine-tuning but often face dimensionality-induced instability in generative tasks. Although kernel-based \citep{Mi2025KerZOOKF} and quantized \citep{zhou-etal-2025-quzo} variants attempt to mitigate gradient variance, estimation noise remains a significant bottleneck. Hybrid approaches such as Addax \citep{li2025addax} combine FO and ZO updates, yet such methods typically rely on static heuristics like sequence length. In contrast, Hi-ZFO employs a principled, sensitivity-guided allocation strategy designed specifically for complex reasoning.

\noindent\textbf{Parameter Efficiency and Adaptive Selection.}
Parameter-efficient fine-tuning (PEFT) \citep{hu2022lora, li-liang-2021-prefix, Houlsby2019ParameterEfficientTL} and adaptive selection methods \citep{huang2024towards} reduce memory overhead by freezing substantial portions of the model backbone. Our work extends the conventional update-or-freeze dichotomy by repurposing inactive layers as sources of beneficial stochasticity via ZO optimization. By dynamically partitioning parameters, the proposed framework maintains gradient precision in critical layers while leveraging ZO updates to facilitate exploration in the remainder. A detailed literature review is provided in Appendix~\ref{app:related_work}.

\section{Method}
\begin{figure*}[t]
    \centering
    \includegraphics[width=0.75\linewidth]{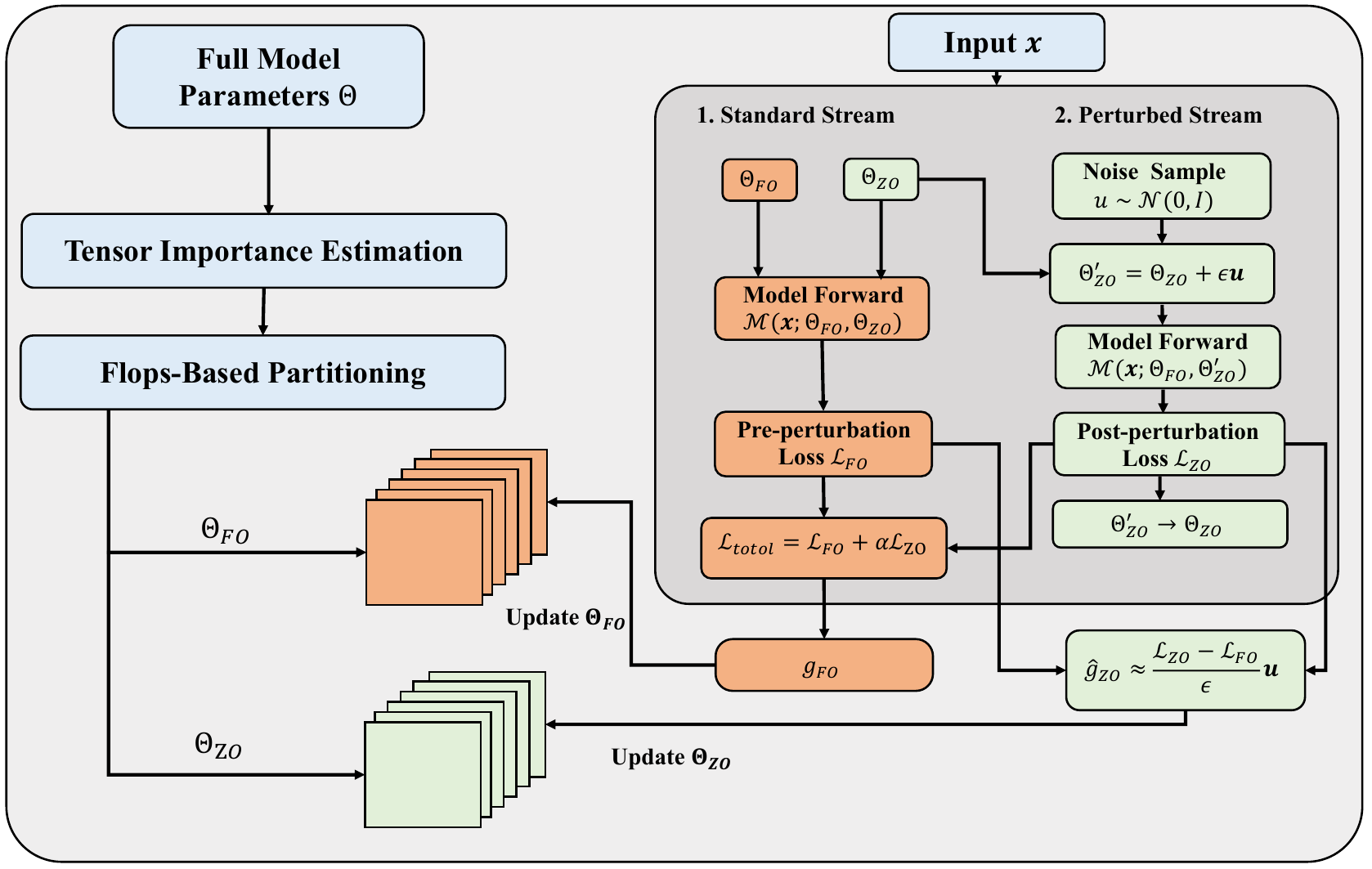}
    \caption{Overview of the Hi-ZFO framework. (a) Partitioning: Splitting $\Theta$ based on sensitivity and cost. (b) Dual-Stream Forward: Evaluating $\mathcal{L}_{\text{FO}}$ and $\mathcal{L}_{\text{ZO}}$ via in-place perturbation. (c) Reset \& Update: Restoring $\Theta_{\text{ZO}}$ using fixed seeds for coordinated FO/ZO updates.}
    \label{fig:model_overview}
\end{figure*}
\subsection{Overview}
Hi-ZFO is a hybrid framework that synergizes First-Order and Zeroth-Order optimization to maximize parameter efficiency under constrained computational resources. 
As illustrated in Figure~\ref{fig:model_overview}, the proposed methodology utilizes a FLOPs-aware criterion to estimate parameter importance, which facilitates the partitioning of model parameters into two distinct sets, $\Theta_{FO}$ and $\Theta_{ZO}$. 
The optimization process is executed through two coupled streams, whereby a standard stream provides accurate gradient signals for critical parameters while a Perturbed Stream drives exploration-based updates for computation-intensive non-critical layers.

\subsection{Cost-Aware Parameter Partitioning}
\label{sec:partitioning}
To optimize resource allocation, we adopt the selection framework established by GreenTrainer \citep{huang2024towards}, which evaluates parameter utility based on loss sensitivity relative to backpropagation cost. Because lower layers require full error propagation, they incur significantly higher FLOPs than upper-layer counterparts. Following the methodology of \cite{huang2024towards}, we utilize Dynamic Programming (DP) to identify a parameter subset ($\Theta_{FO}$) that maximizes cumulative sensitivity within a fixed FLOPs budget, with mathematical details provided in Appendix~\ref{app:importance_selection}.

While the DP solution typically prioritizes FO updates for upper layers, Hi-ZFO departs from the static freezing approach of GreenTrainer by assigning computation-intensive bottom layers to Zeroth-Order optimization ($\Theta_{ZO}$). Since ZO updates rely solely on forward passes, such high-cost layers can be adapted without the prohibitive overhead of backpropagation. This transition effectively repurposes the cost-saving mechanism of GreenTrainer into a source of beneficial exploration, allowing the model to adapt low-level representations while maintaining overall computational efficiency.

\subsection{Dual-Stream Forward Propagation}
\label{sec:dual_stream}
Upon completing parameter partitioning, the optimization procedure executes a sequential dual-stream forward propagation for each input batch $x$. The process commences by generating a perturbed parameter set $\Theta'_{ZO} = \Theta_{ZO} + \epsilon \mathbf{u}$ via a random noise vector $\mathbf{u} \sim \mathcal{N}(\mathbf{0}, \mathbf{I})$ scaled by a smoothing parameter $\epsilon$. 

The dual-stream execution involves two consecutive stages. Initially, the standard stream processes the input using unperturbed parameters $(\Theta_{FO}, \Theta_{ZO})$ to obtain the baseline loss $\mathcal{L}_{FO} = \mathcal{M}(x; \Theta_{FO}, \Theta_{ZO})$. Such a pass facilitates the extraction of essential first-order gradient signals for $\Theta_{FO}$ and establishes a clean reference for subsequent estimation. Subsequently, the perturbed stream performs a second forward pass using unperturbed FO parameters alongside perturbed ZO parameters to yield the post-perturbation loss $\mathcal{L}_{ZO} = \mathcal{M}(x; \Theta_{FO}, \Theta'_{ZO})$. The resulting loss value quantifies the sensitivity of the loss landscape to structured perturbations within the $\Theta_{ZO}$ subspace.

\subsection{Hybrid Optimization}
\label{sec:optimization}
Hi-ZFO utilizes a loss-level coupling to unify FO and ZO optimization within a single objective function, $\mathcal{L}_{total} = \mathcal{L}_{FO} + \alpha \mathcal{L}_{ZO}$, where the hyperparameter $\alpha$ regulates the contribution of the exploration signal. 

Under this unified objective, parameters within $\Theta_{FO}$ undergo standard backpropagation according to the update rule
\begin{equation}
    \theta_{FO} \leftarrow \theta_{FO} - \eta_{FO} \nabla_{\theta_{FO}} \mathcal{L}_{total}.
\end{equation}
Integrating the perturbed loss into $\mathcal{L}_{total}$ explicitly conditions FO optimization on variations within $\Theta_{ZO}$. Such a mechanism prevents upper layers from disregarding shifts in lower layers and encourages the development of solutions robust to representation changes.

Simultaneously, parameters in $\Theta_{\text{ZO}}$ utilize a gradient estimator derived from the finite difference between the standard and perturbed streams. 
To minimize memory overhead, we perform an in-place perturbation by fixing a random seed $s$ to sample the vector $\mathbf{u}$ and setting $\Theta'_{\text{ZO}} \leftarrow \Theta_{\text{ZO}} + \epsilon \mathbf{u}$. 
Immediately after evaluating the perturbed loss $\mathcal{L}_{\text{ZO}}$, the original parameters are recovered in-place by regenerating the same $\mathbf{u}$ via the fixed seed $s$ and computing $\Theta_{\text{ZO}} \leftarrow \Theta'_{\text{ZO}} - \epsilon \mathbf{u}$. 
By leveraging $\mathcal{L}_{\text{FO}}$ as a baseline, the descent direction is calculated as:
\begin{equation}
    \hat{\mathbf{g}}_{\text{ZO}} = \frac{\mathcal{L}_{\text{ZO}} - \mathcal{L}_{\text{FO}}}{\epsilon} \mathbf{u}.
\end{equation}
The resulting gradient estimate is subsequently applied to the restored original parameters $\Theta_{\text{ZO}}$ via the expression:
\begin{equation}
    \Theta_{\text{ZO}} \leftarrow \Theta_{\text{ZO}} - \eta_{\text{ZO}} \hat{\mathbf{g}}_{\text{ZO}}.
\end{equation}
Through such coordinated updates and in-place recovery, Hi-ZFO ensures that FO precision and ZO exploration operate in a mutually aware manner, ultimately stabilizing the optimization trajectory in complex generative landscapes.

\begin{algorithm}[t]
\caption{Hi-ZFO Optimization Algorithm}
\label{alg:Hi-ZFO}
\begin{algorithmic}[1]
\REQUIRE Parameters $\Theta$; Dataset $\mathcal{D}$; FLOPs budget $\mathcal{B}$; Rates $\eta_{\text{FO}}, \eta_{\text{ZO}}$; Scalars $\epsilon, \alpha$; Random seed $s$.
\ENSURE Optimized $\Theta^*$.

\STATE \textbf{Partitioning:} Compute sensitivity $I_k$ and cost $C_k$ for all tensors.
\STATE Solve DP to partition $\Theta \rightarrow \Theta_{\text{FO}}$ and $\Theta_{\text{ZO}}$ s.t. cost $\le \mathcal{B}$.

\FOR{epoch $=1$ to $E$}
    \FOR{batch $(x, y)$ in $\mathcal{D}$}
        \STATE $\mathcal{L}_{\text{FO}} \leftarrow \mathcal{L}(x, y; \Theta_{\text{FO}}, \Theta_{\text{ZO}})$ 
        \STATE Sample $\mathbf{u} \sim \mathcal{N}(\mathbf{0}, \mathbf{I})$ via seed $s$ and set $\Theta'_{\text{ZO}} \leftarrow \Theta_{\text{ZO}} + \epsilon \mathbf{u}$. 

        \STATE $\mathcal{L}_{\text{ZO}} \leftarrow \mathcal{L}(x, y; \Theta_{\text{FO}}, \Theta'_{\text{ZO}})$ 
        
        \STATE $\Theta_{\text{ZO}} \leftarrow \Theta'_{\text{ZO}} - \epsilon \mathbf{u}$ 

        \STATE \textbf{Update $\Theta_{\text{FO}}$ (via Backpropagation):}
        \STATE $\mathcal{L}_{\text{total}} \leftarrow \mathcal{L}_{\text{FO}} + \alpha \mathcal{L}_{\text{ZO}}$
        \STATE $\Theta_{\text{FO}} \leftarrow \Theta_{\text{FO}} - \eta_{\text{FO}} \nabla_{\Theta_{\text{FO}}} \mathcal{L}_{\text{total}}$

        \STATE \textbf{Update $\Theta_{\text{ZO}}$ (via Gradient Estimation):}
        \STATE $\hat{\mathbf{g}}_{\text{ZO}} \leftarrow \frac{\mathcal{L}_{\text{ZO}} - \mathcal{L}_{\text{FO}}}{\epsilon} \mathbf{u}$ 
        \STATE $\Theta_{\text{ZO}} \leftarrow \Theta_{\text{ZO}} - \eta_{\text{ZO}} \hat{\mathbf{g}}_{\text{ZO}}$
    \ENDFOR
\ENDFOR
\RETURN $\Theta_{\text{FO}} \cup \Theta_{\text{ZO}}$
\end{algorithmic}
\end{algorithm}

\subsection{Theoretical Justification}
\label{sec:theory}

To validate the stability of Hi-ZFO, we analyze its convergence properties under standard $L$-smoothness and bounded variance assumptions. The detailed mathematical derivation is provided in Appendix~\ref{app:proofs}. 

\begin{theorem}[Convergence of Hi-ZFO]
\label{thm:convergence}
    Let $\Delta = \mathcal{L}(\theta_0) - \mathcal{L}^*$ denote the initial optimality gap and $\Sigma^2 = \sigma_{FO}^2 + \alpha^2 \sigma_{ZO}^2$ represent the composite variance. By setting the learning rate $\eta = \frac{1}{\sqrt{T}}$ and the ZO smoothing parameter $\mu = \frac{1}{\sqrt{d_{ZO} T}}$, the minimum expected squared gradient norm is bounded by:
    \begin{equation}
    \begin{split}
        \min_{t \in [0, T-1]} \mathbb{E}[\|\nabla \mathcal{L}(\theta_t)\|^2] \leq & \frac{4\Delta + 2L\Sigma^2}{\sqrt{T}} \\
        & + \mathcal{O}\left(\frac{d_{ZO}}{T}\right).
    \end{split}
    \end{equation}
    Consequently, as $T \to \infty$, the algorithm achieves the standard non-convex convergence rate:
    \begin{equation}
        \min_{t \in [0, T-1]} \mathbb{E}[\|\nabla \mathcal{L}(\theta_t)\|^2] \leq \mathcal{O}\left( \frac{1}{\sqrt{T}} \right).
    \end{equation}
\end{theorem}

The convergence behavior is driven by two components. The $\mathcal{O}(1/\sqrt{T})$ term arises from the stochastic variance and optimization gap, confirming that Hi-ZFO matches the asymptotic rate of standard first-order methods. The $\mathcal{O}(d_{ZO}/T)$ term corresponds to the squared smoothing bias, denoted as $2\delta_\mu$ in the derivation (see Appendix \ref{app:proofs}), where $\delta_\mu \propto \mu^2 d_{ZO}^2$. Crucially, this bias term scales linearly with the partial dimension $d_{ZO}$ rather than the full parameter dimension $d$. 

\section{Experiments}
\subsection{Experimental Setup}
\noindent\textbf{Models and Tasks.}
We evaluate Hi-ZFO on a comprehensive suite of LLMs, including OPT \citep{DBLP:journals/corr/abs-2205-01068}, BLOOM \citep{Scao2022BLOOMA1}, Qwen2 \cite{qwen2}, and Qwen2.5-Instruct \citep{qwen2.5}. The assessment spans standard NLP benchmarks, specifically abstractive summarization on SciTLDR \citep{cachola-etal-2020-tldr} and DialogSum \citep{chen-etal-2021-dialogsum} alongside generative QA on WebQuestions \citep{berant-etal-2013-semantic}. Furthermore, we evaluate complex reasoning capabilities on GSM8K \citep{DBLP:journals/corr/abs-2110-14168}, HumanEval \citep{Chen2021EvaluatingLL}, and the Math500 subset of MATH \citep{hendrycks2021measuring}. Such an evaluation focuses exclusively on generative tasks to avoid the metric inflation often observed in simple classification.

\noindent\textbf{Baselines.}
We evaluate Hi-ZFO against a comprehensive suite of baselines, including standard first-order (FO) methods and parameter-efficient fine-tuning (PEFT) techniques such as LoRA \citep{hu2022lora}, GreenTrainer (GT-0.7) \citep{huang2024towards}, and Prefix-Tuning \citep{li-liang-2021-prefix}. We also compare it with zeroth-order (ZO) methods like MeZO \citep{malladi2023mezo} and the hybrid framework Bilevel-ZOFO \citep{shirkavand2025bilevel}. Addax is excluded from this comparison as it defaults to standard first-order optimization in long-sequence scenarios. Detailed formal definitions for all baselines are provided in Appendix~\ref{app:baselines}.

\subsection{Implementation Details}
Experiments conducted on NVIDIA H100 GPUs utilize the AdamW optimizer with decoupled learning rates of $2\times10^{-5}$ for FO and $2\times10^{-6}$ for ZO updates. Hyperparameters for reasoning tasks strictly follow the configurations established in Bilevel-ZOFO \citep{shirkavand2025bilevel}, while standard NLP tasks align with the batch and sequence constraints of GreenTrainer \citep{huang2024towards} to ensure a fair comparison.  In addition, $\epsilon$ is set to $0.001$ and $\alpha$ to $0.1$. Detailed training configurations and the utilized prompt templates are provided in Appendix~\ref{app:training_details} and Appendix~\ref{app:prompts}, respectively.

\begin{table*}[ht]
	\centering
	{\fontsize{7}{8}\selectfont
		\begin{tabular}{lrrrrrrrrr}
			\toprule
			\multirow{2}{*}{\makecell{\textbf{\# Model} \\ \textbf{\& Method}}} & \multicolumn{3}{c}{\textbf{SciTLDR}} & \multicolumn{3}{c}{\textbf{DialogSum}} & \multicolumn{3}{c}{\textbf{WebQuestion}} \\
			\cmidrule(lr){2-4} \cmidrule(lr){5-7} \cmidrule(lr){8-10}
			& \textbf{Mem. (GB)} & \textbf{Time (h)} & \textbf{R1/R2/RL} & \textbf{Mem. (GB)} & \textbf{Time (h)} & \textbf{R1/R2/RL} & \textbf{Mem. (GB)} & \textbf{Time (h)} & \textbf{Acc.}\\
			\midrule
			\rowcolor{gray!25}
			\vspace{0.05in}
			\textbf{OPT-2.7B} \\
			Full FT        & 55.1  & 0.92  & 32.9/14.9/27.1   & 55.1  & 5.5  & 23.6/9.5/18.8 & 54.3 & 0.72 & 29.1  \\
			FT-Top2        & 29.9 & 0.61  & 9.1/4.0/7.6   & 32.7 & 3.8  & 20.8/7.9/17.5 & 26.4 & 0.52 & 23.8   \\
			Prefix-T        & 31.7 & 0.58  & 7.6/0.4/6.1   & 34.0  & 3.7  & 13.4/3.3/10.9  & 27.5 & 0.62 & 25.1 \\
			LoRA        & 29.8  & 0.59   & 28.2/12.1/21.0   & 35.6 & 3.6 & 23.8/9.5/18.8  & 25.1 & 0.5 & 20.0 \\
            MeZO        & \textbf{13.5}  & 2.50  & 6.2/1.5/5.8    & \textbf{13.5} & 12.5  & 10.5/2.1/8.4   & \textbf{13.2} & 1.80 & 12.4 \\
            Bievel-ZOFO & 49.5 & 0.55  & 33.2/15.1/27.4   & 49.5 & 3.5  & 24.1/9.8/19.2 & 48.8 & 0.45 & 29.9 \\
			GT-0.7        & 53.0 & 0.68   & 33.1/15.2/27.6   & 53.0 & 3.9  & 23.4/9.5/18.8  & 51.7 & 0.42 & 30.2 \\ \midrule
            Hi-ZFO      & 45.1 & \textbf{0.42}   & \textbf{35.6/16.7/29.1}   & 45.1 & \textbf{2.5}  & \textbf{26.7/11.0/21.4} & 35.2 & \textbf{0.33} & \textbf{32.4} \\
			\midrule
			\rowcolor{gray!25}
			\vspace{0.05in}
			\textbf{BLOOMZ-3B} \\
			Full FT        & 68.2  & 1.0  & 28.3/12.1/22.5   & 68.2  & 6.5  & 34.1/13.8/27.4  & 60.9 & 0.91 & 21.7 \\
			FT-Top2        & 29.7 & 0.75 & 23.7/8.8/18.8   & 30.8 & 4.6  & 22.1/8.5/17.8   & 27.6 & 0.62 & 18.1 \\
			Prefix-T        & 33.6  & 0.68  & 6.5/2.2/5.5   & 36.7 & 4.2  & 29.6/9.4/24.9  & 29.6 & 0.72 & 15.2 \\
			LoRA        & 29.2  & 0.69  & 27.4/11.7/21.8   & 32.1 & 4.3   & 35.4/14.3/28.6  & 24.8 & 0.51 & 17.4 \\
            MeZO        & \textbf{14.8}  & 2.80  & 5.1/1.2/4.5    & \textbf{14.8} & 14.2  & 12.1/3.5/9.8   & \textbf{14.2} & 2.10 & 10.5 \\
            Bilevel-ZOFO & 54.5 & 0.85  & 28.4/12.3/22.6   & 54.5 & 5.2 & 36.0/14.1/29.5 & 55.2 & 0.75 & 22.6 \\
			GT-0.7        & 57.8 & 0.74  & 28.0/12.2/22.4   & 55.5 & 4.3 & 36.4/13.7/29.2  & 51.2 & 0.65 & 22.7 \\ \midrule
            Hi-ZFO       & 47.9 & \textbf{0.53} & \textbf{29.8/13.1/23.2}   & 47.9 & \textbf{3.2}  & \textbf{38.9/15.9/31.5} & 46.2& \textbf{0.41} & \textbf{24.9}  \\
			\midrule
            \rowcolor{gray!25}
			\vspace{0.05in}
			\textbf{Qwen2.5-3B-Instruct} \\
			Full FT        & 65.4  & 1.20  & 22.1/6.7/17.4   & 65.4  & 7.1  & 21.2/7.2/18.0   & 60.7 & 0.96 & 27.6\\
			FT-Top2        & 33.5 & 0.93 & 16.6/3.4/13.2  & 33.5 & 5.9  & 20.9/6.9/17.8   & 29.4 & 0.72 & 23.7 \\
			Prefix-T        & 31.4  & 0.88  & 20.6/6.8/15.6   & 31.4 & 6.2  & 18.4/5.7/15.5  & 28.7 & 0.79 & 26.7 \\
			LoRA        & 25.6  & 0.65  & 21.7/7.4/16.5   & 25.6 & 5.3   & 22.4/8.6/18.4  & 23.1 & 0.62 & 27.1 \\
            MeZO        & \textbf{14.5}  & 3.50  & 4.2/0.8/3.5    & \textbf{14.5} & 16.5  & 8.5/1.9/6.2    & \textbf{14.0} & 2.50 & 14.2 \\
            Bilevel-ZOFO & 58.2 & 0.95  & 23.5/8.8/18.5   & 58.2 & 5.8  & 22.6/8.9/18.5  & 55.4 & 0.82 & 29.5 \\
			GT-0.7        & 51.3 & 0.84  & 23.6/8.7/18.6  & 51.3 & 6.1 & 22.9/8.9/18.9  & 47.3 & 0.78 & 29.8 \\ \midrule
            Hi-ZFO        & 46.3 & \textbf{0.49}  & \textbf{26.4/10.6/20.5}   & 46.3 & \textbf{4.2}  & \textbf{24.7/10.5/19.8} & 42.5& \textbf{0.47} & \textbf{32.1}  \\
            
			\bottomrule
	\end{tabular}}
	\caption{Performance comparison of different LLMs on SciTLDR, DialogSum, and WebQuestion.}
	\vspace{-0.15in}
	\label{tab:cost_accuracy}
\end{table*}

\subsection{Model Performance}
Table \ref{tab:cost_accuracy} provides a detailed comparison of training costs and downstream performance between Hi-ZFO and several prominent baselines, including Full Fine-Tuning (Full FT), various PEFT methods, and mixed-gradient approaches across three distinct LLM architectures. The reported metrics demonstrate that Hi-ZFO consistently achieves state-of-the-art results, surpassing both Full FT and competitive baselines such as Bilevel-ZOFO and GT-0.7 in most experimental settings. On the DialogSum task using BLOOMZ-3B, for instance, Hi-ZFO obtains a ROUGE-1 score of 38.9, which markedly outperforms the scores of 34.1 and 36.0 achieved by Full FT and Bilevel-ZOFO, respectively. 

Furthermore, Hi-ZFO demonstrates superior time efficiency compared to other high-performance methodologies. Such a speed advantage originates from the rapid convergence of the proposed framework, which reaches peak performance within only three training epochs. In contrast, baseline methods typically require five epochs to converge under their standard configurations, with further training for Hi-ZFO often leading to overfitting. Consequently, the proposed approach substantially reduces computational wall-clock time, such as decreasing the duration from 1.20 hours for Full FT to 0.49 hours on Qwen2.5/SciTLDR. By maintaining a memory footprint lower than Full FT and more efficient than existing mixed-gradient approaches, Hi-ZFO strikes an optimal balance between resource consumption and model accuracy.

\subsection{Performance on Mathematical Reasoning and Coding Tasks.}
Evaluation on reasoning and coding benchmarks using Qwen2-7B reveals a distinct advantage for Hi-ZFO, as detailed in Table~\ref{tab:qwen2_math_code}. Notably, the pure Zeroth-Order baseline (MeZO) collapses on GSM8K, falling from a zero-shot score of 0.420 to 0.329. Such performance degradation highlights the necessity of gradient guidance for complex reasoning, a requirement Hi-ZFO meets with exceptional robustness. Specifically, the proposed framework outperforms Full Fine-Tuning (FT) on both GSM8K (0.810 vs. 0.773) and HumanEval (0.564 vs. 0.505), while also surpassing the mixed-gradient baseline Bilevel-ZOFO (0.543). Although Full FT maintains a marginal lead on Math500 (0.370 vs. 0.350), Hi-ZFO remains competitive and significantly exceeds parameter-efficient methods like LoRA (0.280). The empirical results suggest that the hierarchical strategy acts as an effective regularizer, preserving intrinsic reasoning capabilities more successfully than unconstrained parameter updates.
\begin{table*}[ht]
    \centering
    \vspace{0.1in}
    {\small
    \begin{tabular}{lccc}
        \toprule
        \textbf{Method} & \textbf{GSM8K} (Acc) & \textbf{Math500} (Acc) & \textbf{HumanEval} (pass@1) \\
        \midrule
        Zero-shot & 0.420 & 0.180 & 0.476 \\
        FT (Full Fine-Tuning) & 0.773 & \textbf{0.370} & 0.505 \\
        LoRA          & 0.727 & 0.280 & 0.518 \\
        GT-0.7  & 0.743 & 0.320 & 0.513 \\
        \midrule
        MeZO          & 0.329 & 0.050 & 0.110 \\
        Bilevel-ZOFO  & 0.762 & 0.310 & 0.543 \\
        \textbf{Hi-ZFO} & \textbf{0.810} & 0.350 & \textbf{0.564} \\
        \bottomrule
    \end{tabular}
    }
    \caption{Performance comparison of Qwen2-7B on mathematical and coding tasks. We report Accuracy for GSM8K and Math500, and pass@1 for HumanEval.}
    \label{tab:qwen2_math_code}
\end{table*}

\section{Analysis}
\subsection{Scalability and Efficiency across Model Sizes}
Table \ref{tab:complexity} provides a comprehensive assessment of scalability across model sizes ranging from 350M to 13B parameters. The empirical results consistently indicate that Hi-ZFO achieves a superior trade-off between computational efficiency and downstream accuracy compared to both Full Fine-Tuning and specialized baselines. A significant performance discrepancy emerges when comparing Hi-ZFO to GreenTrainer (GT-0.7), which underscores the fundamental constraints of static parameter freezing. While GT-0.7 minimizes costs by entirely bypassing non-critical parameters, such an approach restricts the model's ability to undergo necessary representation shifts during adaptation. In contrast, Hi-ZFO reactivates these layers through low-cost zeroth-order updates to recover substantial performance, as evidenced by the 2.8\% improvement on the OPT-13B WebQuestions task where Hi-ZFO reaches 35.9\% relative to the 33.1\% obtained by GT-0.7. 
Crucially, the observed gains in accuracy are achieved without increasing training latency. Hi-ZFO records lower wall-clock time than GT-0.7 across all experimental settings, specifically reducing the duration for OPT-13B on SciTLDR from 3.97 to 2.47 hours. Such findings suggest that incorporating exploration-based updates facilitates more effective navigation of high-dimensional parameter spaces than the strict gradient-based selection employed by previous methodologies.
\begin{table*}[ht]
	\centering
	{\fontsize{7}{8}\selectfont
		\begin{tabular}{lrrrrrrrrr}
			\toprule
			\multirow{2}{*}{\makecell{\textbf{\# Params} \\ \textbf{\& Method}}} & \multicolumn{3}{c}{\textbf{SciTLDR}} & \multicolumn{3}{c}{\textbf{DialogSum}} & \multicolumn{3}{c}{\textbf{WebQuestion}} \\
			\cmidrule(lr){2-4} \cmidrule(lr){5-7} \cmidrule(lr){8-10}
			& \textbf{Mem. (GB)} & \textbf{Time (h)} & \textbf{R1/R2/RL} & \textbf{Mem. (GB)} & \textbf{Time (h)} & \textbf{R1/R2/RL}  & \textbf{Mem. (GB)} & \textbf{Time (h)} & \textbf{Acc.}\\
			\midrule
			\rowcolor{gray!25}
			\vspace{0.05in}
			\textbf{OPT-350M} \\
			Full FT        & 12.4  & 0.15  & 30.9/13.9/25.7   & 12.4  & 0.92  & 23.2/9.0/18.5  & 8.4 & 0.18 & 22.7 \\
			LoRA        & 9.8  & 0.10  & 25.9/10.8/20.3   & 9.8  & 0.65   & 21.5/7.7/17.3   & 4.8 & 0.08 & 7.5 \\ 
	
			GT-0.7        & 11.8  & 0.12 & 30.6/13.5/25.0   & 11.8  & 0.66  & 24.2/9.3/19.3  & 8.0 & 0.13 & 20.6 \\
            Hi-ZFO       & 9.7 & 0.09 & 32.3/14.6/26.3   & 9.7  & 0.5 & 25.2/10.5/20.6 & 5.8 & 0.06 & 21.8 \\
			\midrule
			\rowcolor{gray!25}
			\vspace{0.05in}
			\textbf{OPT-1.3B} \\
			Full FT        & 32.9  & 0.46  & 32.1/14.3/26.4   & 32.9  & 2.9  & 25.4/10.3/20.2  & 28.2 & 0.37 & 27.5 \\
			LoRA        & 19.1  & 0.31   & 28.1/11.9/22.0   & 19.1 & 1.9  & 24.6/9.9/19.4  & 14.6 & 0.24 & 15.6 \\
			GT-0.7        & 31.7  & 0.34  & 31.2/14.2/25.8   & 31.7 & 2.0  & 23.4/9.5/18.8  & 26.5 & 0.25 & 28.0 \\
            Hi-ZFO        & 21.2  & 0.2  & 32.8/14.7/26.9   & 21.2  & 1.2 & 27.7/12.6/22.4 & 23.4 & 0.17 & 30.2 \\
			\midrule
			\rowcolor{gray!25}
			\vspace{0.05in}
			\textbf{OPT-2.7B} \\
			Full FT        & 55.1  & 0.92  & 32.9/14.9/27.1   & 55.1  & 5.5  & 23.6/9.5/18.8 & 54.3 & 0.72 & 29.1  \\
			LoRA        & 29.8  & 0.59   & 28.2/12.1/21.0   & 35.6 & 3.6 & 23.8/9.5/18.8  & 25.1 & 0.5 & 20.0 \\
			GT-0.7        & 53.0 & 0.68   & 33.1/15.2/27.6   & 53.0 & 3.9 & 23.4/9.5/18.8  & 51.7 & 0.42 & 30.2 \\
            Hi-ZFO       & 45.1 & 0.42   & 35.6/16.7/29.1   & 45.1 & 2.5  & 26.7/11.0/21.4 & 35.2 & 0.33 & 32.4 \\
			\midrule
			\rowcolor{gray!25}
			\vspace{0.05in}
			\textbf{OPT-6.7B} \\
			Full FT        & 153.2  & 2.32  & 33.9/16.1/28.9  & 153.2  & 10.5 & 25.6/11.3/20.7     & 138.3 & 2.1 & 31.4 \\
			LoRA        & 72.5 & 1.3  & 28.4/12.3/22.7   & 72.5  & 8.1  & 24.9/10.2/19.4  & 69.7 & 1.3 & 22.7 \\
			
			GT-0.7        & 121.7  & 1.4  & 33.1/15.3/27.7  & 121.7  & 8.8  &  26.8/11.2/21.8  & 104.3 & 1.5 & 32.7 \\
            Hi-ZFO        & 110.8  & 1.1  & 35.8/16.9/29.4   & 104.3  & 7.1 & 27.9/11.9/24.3  & 99.6 & 1.0 & 34.8 \\
            \midrule
			\rowcolor{gray!25}
			\vspace{0.05in}
			\textbf{OPT-13B} \\
			Full FT        &326.3  & 5.01 & 33.1/14.9/27.6   & 326.3  & 25.7  & 26.5/10.9/21.0  & 298.7 & 4.95 & 32.8  \\
			LoRA        & 175.6 & 2.96  & 30.7/13.6/25.2   & 175.6 & 18.2  & 26.9/11.3/21.7  & 124.5 & 2.51 & 25.9 \\
			GT-0.7        & 276.3 & 3.97  & 33.9/15.8/28.2  & 276.3  & 22.4  &  27.2/10.5/23.7 & 188.9 & 2.88  & 33.1 \\
            Hi-ZFO        & 210.6 & 2.47  & 36.1/17.2/29.6   & 210.6  & 15.4 & 28.5/12.4/24.9 & 158.4 & 2.11  & 35.9  \\
			\bottomrule
	\end{tabular}}
	\vspace{-0.1in}
	\caption{Impact of LLM's model size}
	\vspace{-0.1in}
	\label{tab:complexity}
\end{table*}
\subsection{Impact of FLOPs-based Update Ratio ($\rho$).}
We further investigate the trade-off between computational budget and downstream performance by varying the FLOPs-based update ratio $\rho$ from 0.2 to 0.8, as detailed in Table~\ref{tab:opt_ratios_comparison}. The results highlight a critical balance between resource consumption and model accuracy. As expected, increasing $\rho$ results in a monotonic rise in training cost; for the OPT-2.7B model, raising $\rho$ from 0.2 to 0.8 more than doubles the training time (1.62h to 3.67h) and expands memory usage from 19.7 GB to 53.0 GB. In terms of performance, we observe an ``inverted-U'' trajectory. At low ratios ($\rho \le 0.3$), the model suffers from underfitting due to insufficient parameter updates, yielding negligible scores. Performance improves sharply as more computationally significant parameters are included, reaching a peak at $\rho=0.6$, where OPT-1.3B and OPT-2.7B achieve optimal ROUGE-1 scores of 27.7 and 26.7, respectively. Crucially, increasing $\rho$ beyond this point yields diminishing returns; at $\rho=0.7$, accuracy notably degrades (e.g., OPT-1.3B drops to 24.2) while memory overhead surges by approximately 50\% (21.2 GB to 32.8 GB). This indicates that updating the top 60\% of parameters based on FLOPs density captures the essential semantic information required for fine-tuning, while higher ratios introduce unnecessary redundancy and resource costs. Consequently, we adopt $\rho=0.6$ as the optimal configuration for our method.
\begin{table*}[ht]
    \centering
    \small
    {\fontsize{8}{10}\selectfont
        \begin{tabular}{lcccccc}
            \toprule
            \multirow{2}{*}{\textbf{Method}} & \multicolumn{3}{c}{\textbf{OPT-1.3B}} & \multicolumn{3}{c}{\textbf{OPT-2.7B}} \\
            \cmidrule(lr){2-4} \cmidrule(lr){5-7}
            & \textbf{Time (h)} & \textbf{Mem (GB)} & \textbf{R1/R2/RL} & \textbf{Time (h)} & \textbf{Mem (GB)} & \textbf{R1/R2/RL} \\
            \midrule
            $\rho$-0.2  & 0.81 & 10.9 & 0.0/0.0/0.0 & 1.62 & 19.7 & 11.7/1.2/8.8 \\
            $\rho$-0.3  & 0.81 & 10.9 & 4.5/0.1/3.4 & 1.67 & 19.7 & 14.3/2.0/10.4 \\
            $\rho$-0.4  & 0.92 & 12.7 & 12.9/3.0/11.1 & 1.86 & 22.4 & 16.7/5.2/14.5 \\
            $\rho$-0.5  & 1.02 & 19.4 & 20.8/6.7/16.3 & 2.12 & 34.1 & 24.5/10.6/20.5 \\
            $\rho$-0.6  & 1.20 & 21.2 & 27.7/12.6/22.4 & 2.50 & 45.1 & 26.7/11.0/21.4 \\
            $\rho$-0.7  & 1.54& 32.8 & 24.2/9.8/19.3 & 3.14 & 53.0 & 24.2/9.8/19.1 \\
            $\rho$-0.8  & 1.71 & 32.8 & 24.9/10.1/19.7 & 3.67 & 53.0 & 25.0/10.3/19.7 \\
            \bottomrule
    \end{tabular}}

    \caption{Performance and efficiency comparison of different pruning ratios on DialogSum.}
    \label{tab:opt_ratios_comparison}
\end{table*}
\begin{figure}[t]
    \centering
    \includegraphics[width=0.7\linewidth]{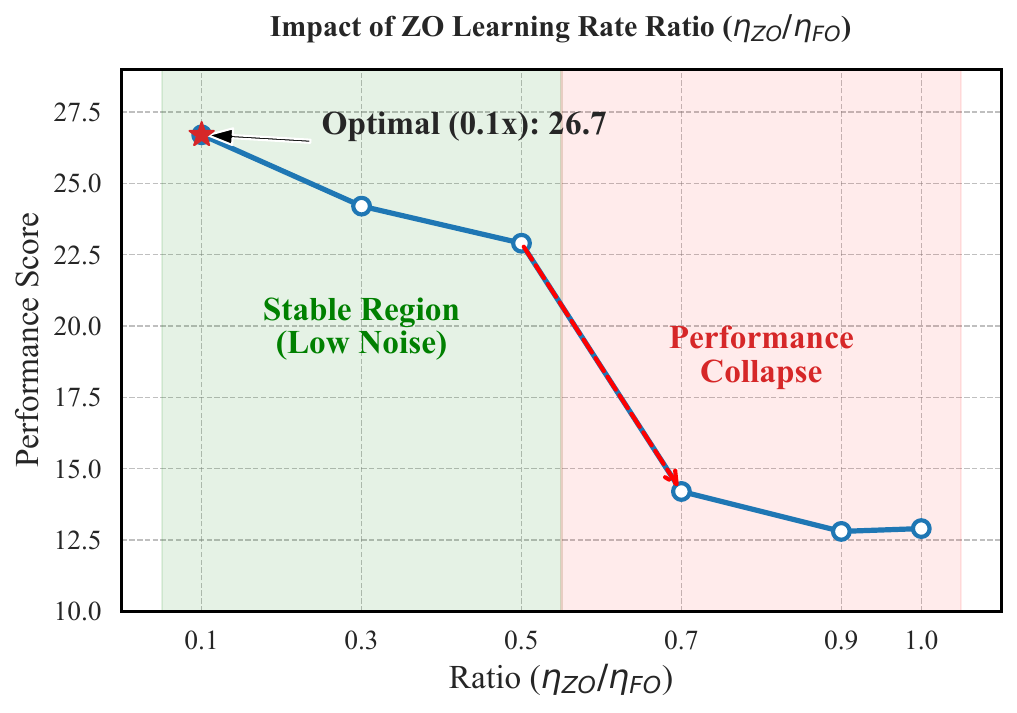}
    \caption{Sensitivity analysis of the learning rate ratio $r = \eta_{\text{ZO}} / \eta_{\text{FO}}$.
    }
    \label{fig:zo_collapse}
\end{figure}
\begin{table*}[ht]
    \vspace{-0.05in}
    \centering
    \small
    {\fontsize{8}{10}\selectfont
        \begin{tabular}{lcccccc}
            \toprule
            \multirow{2}{*}{\textbf{Method ($\alpha$)}} & \multicolumn{3}{c}{\textbf{OPT-2.7B}} & \multicolumn{3}{c}{\textbf{BLOOM-3B}} \\
            \cmidrule(lr){2-4} \cmidrule(lr){5-7}
            & \textbf{SciTLDR} & \textbf{DialogSum} & \textbf{WebQuestion} & \textbf{SciTLDR} & \textbf{DialogSum} & \textbf{WebQuestion} \\
            & \textbf{R1/R2/RL} & \textbf{R1/R2/RL} & \textbf{Acc.} & \textbf{R1/R2/RL} & \textbf{R1/R2/RL} & \textbf{Acc.} \\
            \midrule
             Hi-ZFO & \textbf{35.6/16.7/29.1} & \textbf{26.7/11.0/21.4} & \textbf{32.4} & \textbf{29.8/13.1/23.2} & \textbf{38.9/15.9/31.5} & \textbf{24.9} \\
            \midrule
            w/o ZO-Loss & 32.5/15.0/27.2 & 23.1/9.2/18.8 & 30.2 & 27.5/11.3/22.1 & 36.1/13.2/29.4 & 23.3 \\
            \bottomrule
    \end{tabular}}
    \vspace{-0.05in}
    \caption{Sensitivity analysis of the hyperparameter $\alpha$ (ZO-Loss weight) on OPT-2.7B and BLOOM-3B. The row "w/o ZO-Loss" corresponds to $\alpha=0$.}
    \vspace{-0.15in}
    \label{tab:alpha_ablation}
\end{table*}
\begin{figure}[t]
    \centering
    \includegraphics[width=0.65\linewidth]{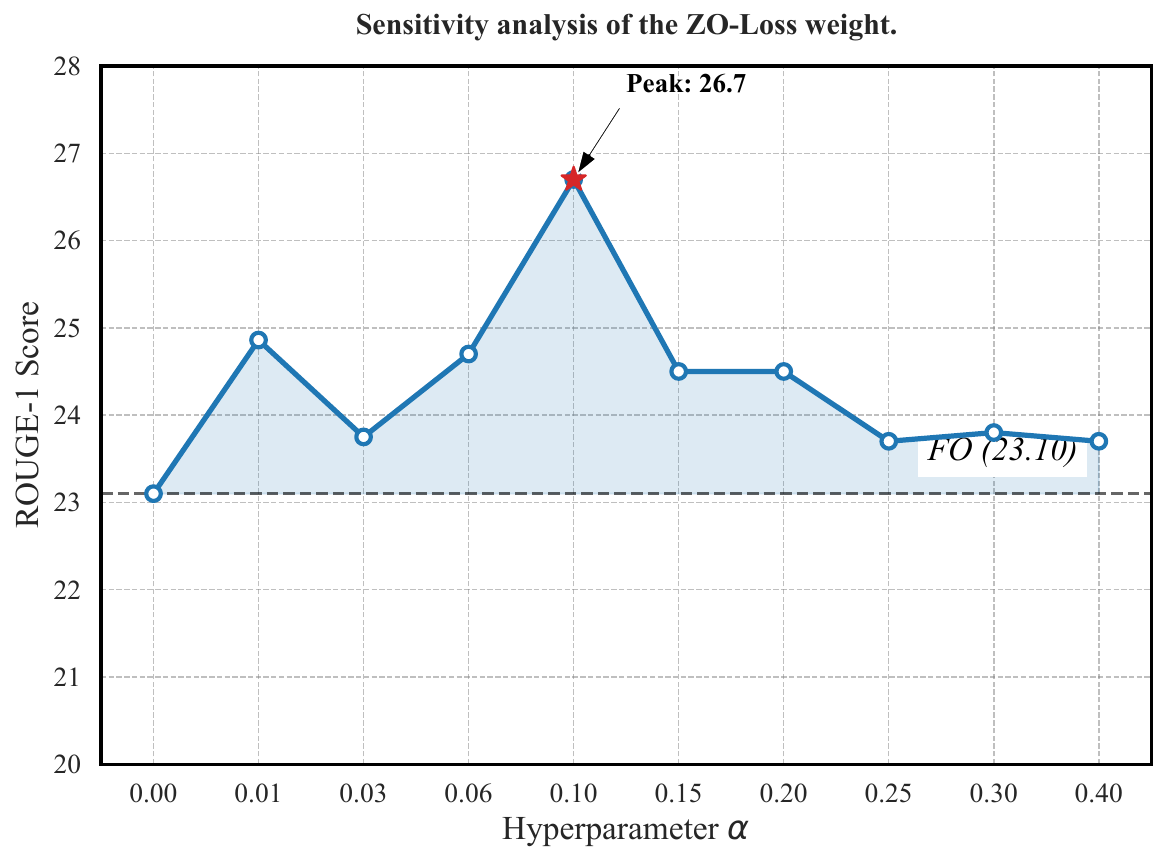}
    \caption{We evaluate the ROUGE-1 score on the DialogSum dataset. Hi-ZFO achieves peak performance at $\alpha=0.1$. 
    }
    \label{fig:alpha_sensitivity}
\end{figure}

\subsection{Impact of ZO Learning Rate Ratio}
To optimize the synergy between first-order (FO) and zeroth-order (ZO) updates, we investigate the sensitivity of the learning rate ratio $r = \eta_{\text{ZO}} / \eta_{\text{FO}}$, as depicted in Figure~\ref{fig:zo_collapse}. The results reveal a clear distinction between a Stable Region and an Unstable Region. Performance peaks (Score: 26.7) at a conservative ratio of $r=0.1$, suggesting that the ZO component functions best as a subtle auxiliary signal. While the model maintains stability within the range $r \in [0.1, 0.5]$, exceeding this threshold precipitates a sharp decline; at $r=0.7$, the score plummets to 14.2. This collapse indicates that the magnitude of ZO updates must be strictly constrained relative to FO updates. When $\eta_{\text{ZO}}$ is too large, the stochastic nature of zeroth-order estimates tends to disrupt the precise trajectory guided by first-order gradients. Consequently, we fix $r=0.1$ to balance exploration with stability.

\subsection{Sensitivity analysis of the ZO-Loss weight $\alpha$.}
To assess the impact of the Zeroth-Order (ZO) regularization term and identify an optimal weighting factor $\alpha$, we conduct a comprehensive ablation and sensitivity analysis, with the findings summarized in Table~\ref{tab:alpha_ablation} and Figure~\ref{fig:alpha_sensitivity}. The results in Table~\ref{tab:alpha_ablation} highlight the indispensability of the ZO loss, as setting $\alpha = 0$ reduces Hi-ZFO to a standard first-order baseline and results in systematic performance attenuation across all evaluated architectures. For instance, removing the ZO component decreases the ROUGE-1 score from 35.6 to 32.5 on OPT-2.7B for the SciTLDR task and from 38.9 to 36.1 on BLOOM-3B for DialogSum.
The sensitivity analysis further characterizes a concave, inverted U-shaped relationship between $\alpha$ and model performance. Accuracy improves as $\alpha$ increases from zero, reaching a peak at $\alpha = 0.1$ with a ROUGE-1 score of 26.7, which substantially outperforms the 23.1 achieved by the first-order baseline. Notably, Hi-ZFO demonstrates significant robustness toward variations in $\alpha$, evidenced by the fact that the performance at $\alpha = 0.4$ remains superior to the pure first-order baseline. Such resilience indicates that the ZO regularization term provides a stable training signal, where the integration of zeroth-order exploration yields consistent benefits over standard fine-tuning regardless of non-optimal weighting.

\section{Conclusion}
In this work, we presented \textbf{Hi-ZFO}, a hybrid optimization framework designed to harmonize the precision of first-order (FO) updates with the exploratory capacity of zeroth-order (ZO) estimation. By implementing a cost-aware parameter partitioning strategy, the proposed framework redefines the utility of ZO optimization, transforming it into a source of structured stochasticity that complements FO learning instead of serving merely as a memory-efficient surrogate. Such a synergy effectively mitigates the overfitting tendencies often associated with full fine-tuning while simultaneously circumventing the convergence instability typical of pure ZO approaches. Empirical evaluations across generative, mathematical, and coding benchmarks demonstrate that Hi-ZFO consistently outperforms both full fine-tuning and established ZO baselines, achieving superior generalization alongside significantly reduced training latency and memory overhead. 
\section*{Limitations}
Although Hi-ZFO demonstrates significant improvements in efficiency and generalization, several avenues for future research remain. 

The first area for potential advancement involves the temporal dynamics of optimization. The current implementation utilizes a stationary parameter partitioning strategy derived from pre-training statistics and maintains a time-invariant ratio between first-order and zeroth-order updates. While such a configuration ensures algorithmic stability, it does not account for the potential evolution of the loss landscape during the fine-tuning process. Because parameter sensitivity often shifts as the model converges, a dynamic mechanism that adaptively recalibrates partition boundaries or modulates exploration magnitude in real-time could theoretically yield superior convergence properties.

A second limitation concerns methodological extensibility, as empirical validation is presently confined to the supervised fine-tuning (SFT) paradigm. With the increasing ascendance of human-centric alignment, the interaction between the proposed dual-stream optimization strategy and preference-based learning frameworks, such as Direct Preference Optimization (DPO), warrants further investigation. 
\bibliography{anthology,custom}
\clearpage

\appendix
\section{Detailed Experimental Settings}
\label{app:details}
\subsection{Baselines Descriptions}
\label{app:baselines}
We compare our proposed method with the following baselines:
\begin{itemize}
    \item Full FT: Standard full-parameter fine-tuning.
    \item FT-Top2: A partial fine-tuning strategy that updates only the top two transformer layers while freezing the rest.
    \item LoRA \citep{hu2022lora}: Low-Rank Adaptation that injects trainable rank-decomposition matrices into transformer layers.
    \item Prefix-T \citep{li-liang-2021-prefix}: Prefix-Tuning, which optimizes continuous task-specific vectors prepended to the keys and values of the attention heads.
    \item GreenTrainer \citep{huang2024towards}: A cost-aware fine-tuning framework that formulates tensor selection as a constrained optimization problem. It identifies the subset of parameters that maximizes sensitivity gain under a fixed FLOPs budget.
    \item MeZO \citep{malladi2023mezo}: The pure zeroth-order optimization method.
    \item Bilevel-ZOFO \citep{shirkavand2025bilevel}: A hybrid framework employing a double-loop optimization strategy. 
\end{itemize}

\subsection{Training Duration and Efficiency}
\label{app:training_details}
Following the standard protocol from GreenTrainer \citep{huang2024towards}, we configure all baseline models with a learning rate of $2\times 10^{-5}$ and batch sizes of either 4 or 32. For our proposed Hi-ZFO, we limit the training duration to 3 epochs, as empirical results indicated that excessive training epochs led to overfitting. Despite this reduced duration, Hi-ZFO exhibits superior data efficiency, achieving competitive or superior performance compared to the baselines.

\subsection{Prompt Templates}
\label{app:prompts}
To ensure fair comparison, we use specific input formatting for each task:
\begin{itemize}
    \item Reasoning \& Coding (GSM8K, MATH, HumanEval):\\
    \texttt{Question: \{Q\} \textbackslash n Answer: \{A\}}
    \item QA (WebQuestions):\\
    \texttt{question: \{q\} </s> answer: \{a\} </s>}
    \item Summarization (SciTLDR, DialogSum):\\
    \texttt{[source seq.] TL;DR:}
\end{itemize}
\section{Extended Related Work}
\label{app:related_work}

\subsection{Advanced Zeroth-Order Methods}
Beyond the seminal MeZO \citep{malladi2023mezo}, several works have addressed ZO limitations. KerZOO \citep{Mi2025KerZOOKF} introduces kernel-informed estimators to reduce gradient variance. Others integrate ZO with low-rank structures (e.g., LoRA) to constrain the search space \citep{chen2025enhancing, 10711229}. Specialized frameworks like QuZO \citep{zhou-etal-2025-quzo} adapt ZO for quantized models. Despite these advances in estimator design, pure ZO methods often struggle with optimization stability in high-dimensional generative tasks.

\subsection{Heuristic Hybrid Strategies}
Hybrid optimization aims to balance FO precision and ZO efficiency. Early benchmarks \citep{zhang2024revisiting} explored simple block-wise update schemes. More complex methods like BiLevel-ZOFO \citep{shirkavand2025bilevel} use gradient fusion for meta-training. Addax \citep{li2025addax} proposes a length-based heuristic, applying FO to long sequences and ZO to short ones. However, these methods often fail to address the intrinsic sensitivity of different model parameters, relying instead on external data attributes or fixed architectural patterns.

\section{Details of Tensor Importance Evaluation and Selection}
\label{app:importance_selection}

\subsection{Mathematical Formulation of Tensor Importance}
The importance of a tensor is estimated as the summation of the importance values of its constituent weights. Since model weights are updated to minimize training loss, the importance of a weight update in a given iteration can be evaluated by the potential increase in loss, $\Delta L = L(w) - L(w + \Delta w)$, if the update were reversed. Given that computing $\Delta L$ for every weight is computationally expensive, we estimate the importance of all weights simultaneously by smoothing the reversal operation and calculating gradients with respect to the updates. 

Let a multiplicative vector $\bm{c} \in [0,1]^M$ denote the reversal operation for all $M$ weights. The loss gradient is expressed as
\begin{equation}\label{eq:soft_undo}
\begin{split}
    -\frac{\partial L(\bm{w} + \bm{c} \odot \Delta \bm{w})}{\partial \bm{c}} & \\
    = -\left. \Delta \bm{w} \odot \frac{\partial L(\bm{u})}{\partial \bm{u}} \right\vert_{\substack{\bm{u}=\bm{w} \\ + \bm{c} \odot \Delta \bm{w}}}
\end{split}
\end{equation}
where $\odot$ denotes element-wise multiplication. When $\bm{c} = \bm{0}$, the resulting expression provides an importance vector across all weights. Because the loss gradient is parameterized by the entire weight set, the weight importance calculated in this manner implicitly incorporates the impact of weight dependencies. Consequently, the importance $I_k$ of a tensor $k$ is calculated as
\begin{align}\label{eq:importance}
I_k &= -\sum\nolimits_{i} \Delta w_i^{(k)} \frac{\partial L}{\partial w_i^{(k)}}.
\end{align}
To improve numerical stability during the tensor selection process, especially when training encounters high variance, all tensor importance values are scaled by their maximum amplitude to prevent potential overflow.

\subsection{Subproblem Decomposition in Dynamic Programming}
Since the tensor selection problem is a nonlinear integer programming problem and thus NP-hard, we utilize Dynamic Programming (DP) to identify an approximate solution. The problem is decomposed into a series of subproblems, $P[k,t]$, which seek to maximize the cumulative importance of selected tensors among the top $k$ layers under a backpropagation FLOPs constraint $t$. The layers are indexed starting from the one closest to the output.

The solution to each subproblem $P[k,t]$ is derived from previously solved smaller subproblems through a defined recurrence relation. If tensor $k$ is not selected, the value of $P[k,t]$ is equivalent to $P[k-1,t]$. Alternatively, if tensor $k$ is selected, the required FLOPs budget must account for both the update of tensor $k$ and the propagation of activation gradients from the nearest selected tensor $k_c$. This requirement is represented by the recurrence equation
\begin{equation} 
\Delta t = t_{dw_k}+\sum\nolimits_{j=k_c}^{k-1}t_{dy_j}.
\label{eq:recursion}
\end{equation}
Because the optimal $k_c$ is unknown in advance, the algorithm backtraces through the subproblem space to explore all possible values of $k_c$. The optimal solution for $P[k,t]$ is identified as the one yielding the highest cumulative importance. This process continues until the global problem $P[N, T_{full}]$ is resolved. The overall time complexity of the DP approach is $O(N^2T_{full})$, which is efficient for typical large language model configurations.

\section{Detailed Theoretical Proofs}
\label{app:proofs}

In this section, we provide the detailed mathematical derivation for the convergence analysis of the Hi-ZFO algorithm.

\subsection{Notations and Problem Setup}
We consider the minimization of a non-convex objective function $\mathcal{L}(\theta)$, where the parameter vector $\theta \in \mathbb{R}^d$ is partitioned into lower layers $\theta_{ZO} \in \mathbb{R}^{d_{ZO}}$ and upper layers $\theta_{FO} \in \mathbb{R}^{d_{FO}}$. The total parameter vector is $\theta = [\theta_{ZO}^\top, \theta_{FO}^\top]^\top$.

The update rule at iteration $t$ is given by:
\begin{equation}
    \theta_{t+1} = \theta_t - \eta \mathbf{v}_t,
\end{equation}
where $\eta > 0$ is the learning rate. The gradient estimator vector $\mathbf{v}_t$ is constructed as:
\begin{equation}
    \mathbf{v}_t = \begin{bmatrix} \alpha \cdot \hat{\mathbf{g}}_{ZO, t} \\ \mathbf{g}_{FO, t} \end{bmatrix},
\end{equation}
where:
\begin{itemize}
    \item $\mathbf{g}_{FO, t}$ is the stochastic First-Order (FO) gradient estimator for the upper layers.
    \item $\hat{\mathbf{g}}_{ZO, t}$ is the Zeroth-Order (ZO) gradient estimator for the lower layers, computed using coordinate-wise smoothing or random vector Gaussian smoothing with smoothing parameter $\mu$.
    \item $\alpha > 0$ is a weighting coefficient balancing the updates.
\end{itemize}

To rigorously analyze the convergence, we define the \textit{effective} total gradient that the algorithm aims to approximate as $\mathbf{G}(\theta_t) = [\alpha \nabla_{\theta_{ZO}}\mathcal{L}, \nabla_{\theta_{FO}}\mathcal{L}]^\top$. We assume the loss scaling aligns such that $\mathbf{v}_t$ is a valid estimator for the descent direction of $\mathcal{L}$.

\subsection{Assumptions}

\begin{assumption}[$L$-Smoothness] \label{as:smooth}
    The loss function $\mathcal{L}(\theta)$ is $L$-smooth, i.e., differentiable and there exists a constant $L > 0$ such that for all $\theta, \theta' \in \mathbb{R}^d$:
    \begin{equation}
    \begin{split}
        \mathcal{L}(\theta') \leq \mathcal{L}(\theta) &+ \langle \nabla \mathcal{L}(\theta), \theta' - \theta \rangle \\
        &+ \frac{L}{2} \|\theta' - \theta\|^2.
    \end{split}
    \end{equation}
\end{assumption}

\begin{assumption}[First-Order Variance] \label{as:fo_variance}
    The FO gradient estimator is unbiased and has bounded variance:
    \begin{align}
        \mathbb{E}[\mathbf{g}_{FO, t}] &= \nabla_{\theta_{FO}} \mathcal{L}(\theta_t), \\
        \mathbb{E}[\|\mathbf{g}_{FO, t} - \nabla_{\theta_{FO}} \mathcal{L}(\theta_t)\|^2] &\leq \sigma_{FO}^2.
    \end{align}
\end{assumption}

\begin{assumption}[Zeroth-Order Properties] \label{as:zo_props}
    The ZO gradient estimator $\hat{\mathbf{g}}_{ZO, t}$ (generated via Gaussian smoothing with parameter $\mu$ and dimension $d_{ZO}$) satisfies:
    \begin{align}
        \|\mathbb{E}[\hat{\mathbf{g}}_{ZO, t}] - \nabla_{\theta_{ZO}} \mathcal{L}(\theta_t)\|^2 &\leq \delta_\mu, \label{eq:zo_bias} \\
        \mathbb{E}[\|\hat{\mathbf{g}}_{ZO, t}\|^2] \leq 2(d_{ZO} + 1) &\|\nabla_{\theta_{ZO}} \mathcal{L}(\theta_t)\|^2 \nonumber \\
        &+ \sigma_{ZO}^2, \label{eq:zo_moment}
    \end{align}
    where $\delta_\mu = \mathcal{O}(\mu^2 d_{ZO}^2)$ is the bias and $\sigma_{ZO}^2$ is the sampling variance.
\end{assumption}

\subsection{Convergence Analysis}

\begin{lemma}[One-step Descent] \label{lemma:descent}
    Under Assumption \ref{as:smooth}, the iteration satisfies:
    \begin{equation}
    \begin{split}
        \mathbb{E}_t[\mathcal{L}(\theta_{t+1})] \leq \mathcal{L}(\theta_t) &- \eta \mathbb{E}_t[\langle \nabla \mathcal{L}(\theta_t), \mathbf{v}_t \rangle] \\
        &+ \frac{L\eta^2}{2} \mathbb{E}_t[\|\mathbf{v}_t\|^2],
    \end{split}
    \end{equation}
    where $\mathbb{E}_t[\cdot]$ denotes the expectation conditioned on $\theta_t$.
\end{lemma}

\subsubsection{Derivation of the Bound}

Let $\nabla \mathcal{L}_t$ be the full gradient. We analyze $\langle \nabla \mathcal{L}(\theta_t), \mathbb{E}_t[\mathbf{v}_t] \rangle$. Focusing on the ZO bias (assuming $\alpha \approx 1$ for simplicity):
\begin{align}
    \langle \nabla \mathcal{L}_t, &\mathbb{E}_t[\mathbf{v}_t] \rangle \nonumber \\
    &= \|\nabla \mathcal{L}_t\|^2 + \langle \nabla \mathcal{L}_t, \mathbb{E}_t[\mathbf{v}_t] - \nabla \mathcal{L}_t \rangle \nonumber \\
    &= \|\nabla \mathcal{L}_t\|^2 + \langle \nabla_{\theta_{ZO}} \mathcal{L}_t, \mathbb{E}[\hat{\mathbf{g}}_{ZO, t}] - \nabla_{\theta_{ZO}} \mathcal{L}_t \rangle.
\end{align}
Using $\langle \mathbf{x}, \mathbf{y} \rangle \geq -\frac{1}{2}\|\mathbf{x}\|^2 - \frac{1}{2}\|\mathbf{y}\|^2$:
\begin{equation} \label{eq:inner_prod_bound}
\begin{split}
    \langle \nabla \mathcal{L}_t, \mathbb{E}_t[\mathbf{v}_t] \rangle \geq \|\nabla \mathcal{L}_t\|^2 &- \frac{1}{2}\|\nabla_{\theta_{ZO}} \mathcal{L}_t\|^2 \\
    &- \frac{1}{2}\|\mathbb{E}[\hat{\mathbf{g}}_{ZO}] - \nabla_{\theta_{ZO}} \mathcal{L}_t\|^2.
\end{split}
\end{equation}
Substituting the bias bound (Eq. \ref{eq:zo_bias}):
\begin{equation}
    \langle \nabla \mathcal{L}_t, \mathbb{E}_t[\mathbf{v}_t] \rangle \geq \frac{1}{2}\|\nabla \mathcal{L}_t\|^2 - \frac{1}{2}\delta_\mu.
\end{equation}
\textit{Note: We used $\|\nabla \mathcal{L}\|^2 - \frac{1}{2}\|\nabla_{ZO}\mathcal{L}\|^2 \geq \frac{1}{2}\|\nabla \mathcal{L}\|^2$.}

We bound $\mathbb{E}_t[\|\mathbf{v}_t\|^2] = \alpha^2 \mathbb{E}[\|\hat{\mathbf{g}}_{ZO}\|^2] + \mathbb{E}[\|\mathbf{g}_{FO}\|^2]$. Using Assumptions \ref{as:fo_variance} and \ref{as:zo_props}:
\begin{align}
    \mathbb{E}_t[\|\mathbf{v}_t\|^2] &\leq \alpha^2 \left( 2(d_{ZO}+1)\|\nabla_{\theta_{ZO}}\mathcal{L}\|^2 + \sigma_{ZO}^2 \right) \nonumber \\
    &\quad + \left( \|\nabla_{\theta_{FO}}\mathcal{L}\|^2 + \sigma_{FO}^2 \right) \nonumber \\
    &\leq C_g \|\nabla \mathcal{L}(\theta_t)\|^2 + \Sigma^2, \label{eq:second_moment}
\end{align}
where $C_g = \max(2\alpha^2(d_{ZO}+1), 1)$ and $\Sigma^2 = \alpha^2 \sigma_{ZO}^2 + \sigma_{FO}^2$.

Substitute Eq. \eqref{eq:inner_prod_bound} and \eqref{eq:second_moment} into Lemma \ref{lemma:descent}:
\begin{align}
    \mathbb{E}_t[\mathcal{L}_{t+1}] &\leq \mathcal{L}_t - \eta \left( \frac{1}{2}\|\nabla \mathcal{L}_t\|^2 - \frac{1}{2}\delta_\mu \right) \nonumber \\
    &\quad + \frac{L\eta^2}{2} (C_g \|\nabla \mathcal{L}_t\|^2 + \Sigma^2) \nonumber \\
    &= \mathcal{L}_t - \frac{\eta}{2} (1 - L\eta C_g) \|\nabla \mathcal{L}_t\|^2 \nonumber \\
    &\quad + \frac{\eta}{2}\delta_\mu + \frac{L\eta^2}{2}\Sigma^2.
\end{align}

Assume $\eta \leq \frac{1}{2L C_g}$, so $1 - L\eta C_g \geq \frac{1}{2}$:
\begin{equation}
    \mathbb{E}_t[\mathcal{L}_{t+1}] \leq \mathcal{L}_t - \frac{\eta}{4} \|\nabla \mathcal{L}_t\|^2 + \frac{\eta}{2}\delta_\mu + \frac{L\eta^2}{2}\Sigma^2.
\end{equation}
Rearranging to isolate the gradient norm:
\begin{equation}
    \|\nabla \mathcal{L}_t\|^2 \leq \frac{4}{\eta} (\mathcal{L}_t - \mathbb{E}_t[\mathcal{L}_{t+1}]) + 2\delta_\mu + 2L\eta \Sigma^2.
\end{equation}
Summing over $t=0 \dots T-1$ and taking expectation:
\begin{equation}
\begin{split}
    \sum_{t=0}^{T-1} \mathbb{E}[\|\nabla \mathcal{L}(\theta_t)\|^2] \leq &\frac{4}{\eta} (\mathcal{L}_0 - \mathcal{L}^*) \\
    &+ 2T\delta_\mu + 2L T \eta \Sigma^2.
\end{split}
\end{equation}
Dividing by $T$:
\begin{equation}
    \frac{1}{T} \sum_{t=0}^{T-1} \mathbb{E}[\|\nabla \mathcal{L}_t\|^2] \leq \frac{4(\mathcal{L}_0 - \mathcal{L}^*)}{\eta T} + 2\delta_\mu + 2L\eta \Sigma^2.
\end{equation}

Setting $\eta = \frac{1}{\sqrt{T}}$ and $\mu = \frac{1}{\sqrt{d_{ZO} T}}$ (recall $\delta_\mu \propto \mu^2 d_{ZO}^2$), we get:
\begin{equation}
    \min_{t} \mathbb{E}[\|\nabla \mathcal{L}(\theta_t)\|^2] \leq \mathcal{O}\left( \frac{1}{\sqrt{T}} \right).
\end{equation}
This concludes the proof.

\end{document}